\documentclass[letterpaper, 10 pt, conference]{ieeeconf}  % Comment this line out if you need a4paper

\IEEEoverridecommandlockouts                              % This command is only needed if 
\usepackage{hyperref}
\usepackage{url}
\usepackage{graphics} % for pdf, bitmapped graphics files
\usepackage{times} % assumes new font selection scheme installed
\usepackage{booktabs}       % professional-quality tables
\usepackage{amsfonts}       % blackboard math symbols
\usepackage{nicefrac}       % compact symbols for 1/2, etc.
\usepackage{microtype}      % microtypography
\usepackage{algorithm,multirow,xcolor}
\usepackage{algorithmicx}
\usepackage{algpseudocode}
\usepackage{mathtools}
\usepackage{graphicx}
\usepackage{cases}
\usepackage{array}
\usepackage{color}
\usepackage{float}
\usepackage{amssymb}
\usepackage{bm}
\usepackage{cite}

\usepackage[skip=5pt]{caption}

\usepackage{wrapfig}
\usepackage{subfigure}
\usepackage{amsmath}
\usepackage{amsthm, amssymb}
\theoremstyle{definition}

\usepackage{soul}

\usepackage{graphicx} 
\usepackage{subfigure}
\usepackage{epstopdf}
\makeatletter

\newcommand{\ych}[1]{\textcolor{red}}
\newcommand{\delete}[1]{\textcolor{blue}}
\makeatother
\title{\LARGE \bf

	GRF-based Predictive Flocking Control with Dynamic Pattern Formation
}

\author{Chenghao Yu$^{*,1}$, Dengyu Zhang$^{*,1}$, and Qingrui Zhang$^{1}$ % <-this % stops a space
\thanks{This work is supported by the National Nature Science Foundation of China under Grant 62103451, in part by Shenzhen Science and Technology Program JCYJ20220530145209021.}
\thanks{$^*$These authors contributed equally to this work. $^{1}$The authors are with the School of Aeronauics and Astronautics, Shenzhen Campus of Sun Yat-sen University, Shenzhen 518107, P.R. China. Correspondence to Qingrui Zhang ({\tt \small zhangqr9@mail.sysu.edu.cn}) }
}

\begin{document}
	
	\maketitle
	\thispagestyle{empty}
	\pagestyle{empty}

	%%%%%%%%%%%%%%%%%%%%%%%%%%%%%%%%%%%%%%%%%%%%%%%%%%%%%%%%%%%%%%%%%%%%%%%%%%%%%%%%
	\begin{abstract}
It is promising but challenging to design flocking control for a robot swarm to autonomously follow changing patterns or shapes in a optimal distributed manner. The optimal flocking control with dynamic pattern formation is, therefore, investigated in this paper. A predictive flocking control algorithm is proposed based on a Gibbs random field (GRF), where bio-inspired potential energies are used to charaterize ``robot-robot'' and ``robot-environment'' interactions. Specialized performance-related energies, \emph{e.g.}, motion smoothness, are introduced in the proposed design to improve the flocking behaviors. The optimal control is obtained by maximizing a posterior distribution of a GRF. A region-based shape control is accomplished for pattern formation in light of a mean shift technique. The proposed algorithm is evaluated via the comparison with two state-of-the-art flocking control methods in an environment with obstacles. Both numerical simulations and real-world experiments are conducted to demonstrate the efficiency of the proposed design.

	\end{abstract}

	%%%%%%%%%%%%%%%%%%%%%%%%%%%%%%%%%%%%%%%%%%%%%%%%%%%%%%%%%%%%%%%%%%%%%%%%%%%%%%%%
	\section{Introduction}

Gregarious animals---from starling birds to bees---are adept at spectacular coordinated flocking behaviors in diverse environments to achieve complex tasks \cite{cavagna2010, Bialek2012PNAS, Beaver2021ARC}. Such interesting phenomena have intrigued researchers from many disciplines for a long time \cite{pimenta2013TR,rubenstein2014,zhang2017JA, schilling2019RAL, yang2022Nature}. In particular, flocking of collective robots has gained extensive attention due to its extraordinary promise in various applications, including agriculture, package delivery, search and rescue,  \emph{etc.} \cite{Hu2021TRO,Zhang2021JGCD,ZhangZheng2022}. For large gatherings of robots, flocking control is commonly designed in a decentralized or distributed fashion, where robots leverage local observations and simple rules for decision-making \cite{Chung2018TRO,Beaver2021ARC}.  However, it is challenging to ensure both safe flocking behaviors and dynamic pattern formation for robots with limited onboard resources.

Most of the existing algorithms are heuristic rule-based methods consisting of multiple bio-inspired meta rules, for instance, \emph{separation} for collision avoidance, \emph{cohesion} for robot congregation, and \emph{alignment} for motion consensus, \emph{etc.}, as Reynolds flocking \cite{Reynolds1987SigC}. The heuristic rule-based flocking control is designed based on either artificial potential fields (APF) \cite{Bennet2010RAS} or certain velocity vectors \cite{Vicsek2012PR}, so it is a reactive method with robots determining their motion by the instantaneous measurements from their proprioceptive and exteroceptive sensors \cite{Vicsek2012PR}. The rule-based flocking is suitable for lightweight platforms with limited resources \cite{Zhou2022SR}, but the emergence of safe flocking is often acquired at the cost of non-smooth and oscillatory motion. Pattern formation is a difficult task for the rule-based flocking. Hence, two questions naturally arose:
\begin{enumerate}
    \item \emph{How could we improve the flocking control performance with respect to more metrics, e.g., motion smoothness?}
    \item \emph{How could we achieve dynamic pattern formation or region-based shape control for robots in flocking?}
\end{enumerate}
    \begin{figure}[!tbbp]
        \centering
        \includegraphics[width=\linewidth]{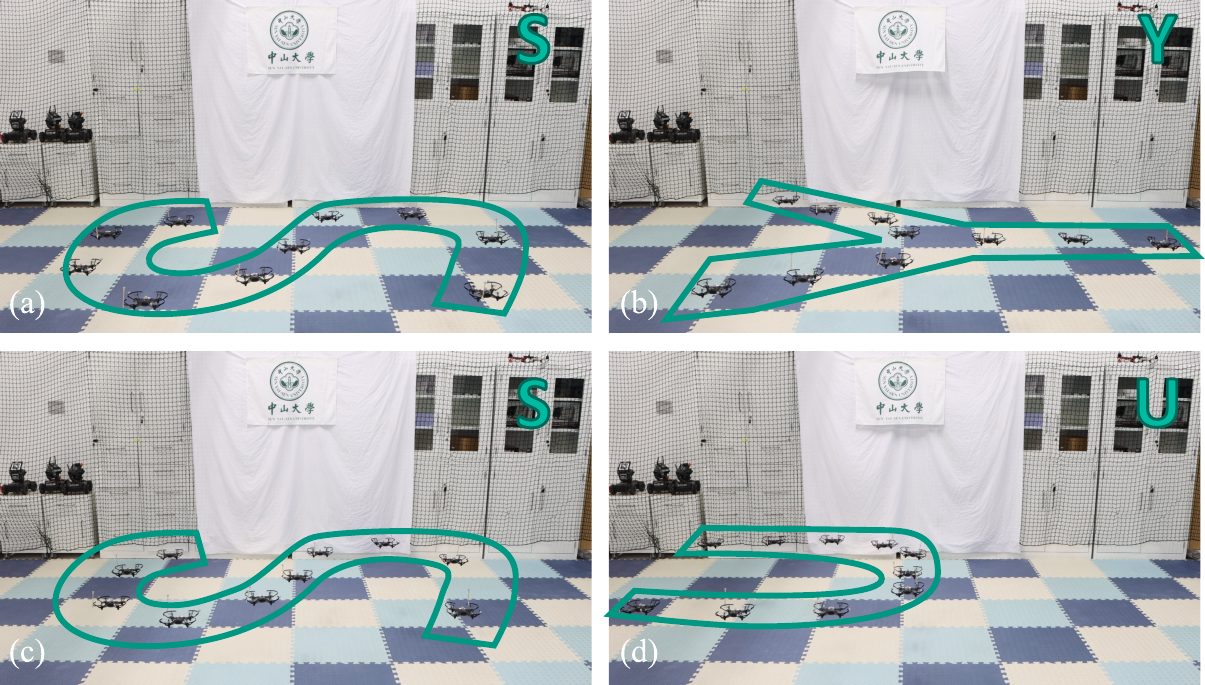}
        \caption{Dynamic pattern formation experiments. Ten UAVs forming ``S", ``Y", ``S", and ``U" patterns in a sequence. }

        \label{fig:real trajectory}
        \vspace{-0.5cm}
    \end{figure}

For the first question above, optimization-based flocking control has been investigated, which solves an optimization problem online in a recursive way \cite{soria2021Nature,zhan_flocking_2013}. The optimization-based flocking control, \emph{e.g.},  model predictive control (MPC), leverages a mathematical model to predict and optimize the future behaviors of robots \cite{lyu_multivehicle_2021}. In traditional MPC, the optimization of one robot depends on the optimization of the neighboring robots, and \emph{vice versa}. Hence, traditional MPC suffers from the decision conflict issue. It is also difficult to determine appropriate costs for the optimization problem. Another solution to the aforementioned first question is based on Gibbs Random Field (GRF) \cite{Xi2006Auto, TAN20102068, Rezeck2021IROS, fernando_online_2021, rezeck2022RAL,zhu_online_2023}. In GRF, a robot swarm is formulated as a collection of random variables at spatial sites \cite{Xi2006Auto, TAN20102068}. The robot-robot and robot-environment interactions are characterized by potential energies that are constructed based on APFs for flocking \cite{fernando_online_2021}. On the basis of the potential energies, a joint probability distribution function is established to represent the spatial flock configuration \cite{fernando_online_2021}. The objective of flocking control would, therefore, be transformed into the inference of the optimal control that maximizes the posterior joint distribution of the GRF at a future time instant. GRF provides a compelling framework that combines bio-inspired methods and optimization-based solutions. Hence, GRF has competence in the development of the optimal robot flocking control. However, existing GRF-based methods failed to take motion-related performance into account, \emph{e.g.}, motion smoothness. More importantly, dynamic pattern formation or region-based shape control is an open issue for the current GRF-based flocking designs.

In this paper, the optimal flocking with dynamic pattern formation problem is investigated based on GRF, which aims to resolve the two aforementioned questions in one framework. The first contribution of this paper is the development of a GRF-based predictive flocking control algorithm, where GRF is used to characterize the interactions among robots and the motion performance of interest. A posterior distribution of the GRF is approximated distributely using the potential energies computed based on predicted forthcoming robot states. The approximation is performed online in an iterative way in light of the mean-field theory. The usage of predicted information can potentially enhance the flocking performance. The optimal flocking control is, therefore, obtained for each robot in a distributed fashion by maximizing the approximated a posterior distribution of the GRF. The competence of the GRF-based control is demonstrated via the comparison with the state-of-the-art methods in \cite{olfati-saber_flocking_2006} and \cite{Vasarhelyi2018SR}, respectively. 

The second contribution of this paper is the design of several potential energies that are helpful for the improvement of the flocking performance. For example, an alignment potential energy improves motion consistency; an obstacle avoidance potential energy makes it easier for a robot swarm to pass through obstacles; and a motion smoothness energy ensures that the magnitude and direction of the acceleration cannot be significantly changed, thus improving motion smoothness. Numerical simulations were performed to illustrate the efficiency of the new designs.

The last contribution of this paper is the introduction of an exploration potential energy based on the mean shift clustering \cite{Yizong1995PAMI}. With the exploration potential energy, a robot swarm has the capability of dynamic pattern formation. It means a robot swarm could track an arbitrary region-based shape. Both simulation and experiments are conducted to demonstrate the efficiency of the dynamic pattern formation of the proposed design.

    The rest of the paper is organized as follows. Section \ref{sec:related} describes the related works of flocking control. In Section \ref{sec:Prob}, the problem formulation is provided. Section \ref{sec:appr} describes the proposed flocking control framework. Section \ref{sec:exp} presents the simulation and experimental results. Conclusion remarks are given in Section \ref{sec:Conclusion}.
	
	\section{Related works\label{sec:related}}
    Different flocking control algorithms have been developed based on bio-inspired heuristic rules designed directly or indirectly in light of APFs. In \cite{olfati-saber_flocking_2006}, Olfati-Saber \emph{et. al.} proposed three distributed flocking control algorithms, each of which fundamentally follows Reynolds flocking rules via potential functions. The concept of virtual intelligent robots ($\beta$-agents) is introduced, which plays a significant role in obstacle avoidance. In \cite{Vasarhelyi2018SR}, V\'{a}s\'{a}rhelyi \emph{et. al.} proposed a distributed heuristic flocking control algorithm for robots with first-order dynamics, where the algorithm parameters are optimized using an evolutionary algorithm. The parameter optimization process in \cite{Vasarhelyi2018SR} is computationally expensive, and the obstacle avoidance performance would degrade significantly in a complex cluttered environment. 
    In \cite{cucker2007TAC}, the well-known Cucker-Smale (C-S) model is proposed by assuming that every robot adjusts its velocity using a weighted average of the differences of its velocity with those of the other robots, which has improved convergence.  Albani \emph{et. al.} evaluate the APF-based flocking control in three dimensions in real-world environments \cite{albani2022ICRA}. However, the heuristic rule-based flocking control is reactive, which lacks optimality. Physical constraints are satisfied at the cost of laborious parameter tuning \cite{Vasarhelyi2018SR}. Dynamic pattern formation is a non-trivial task for rule-based methods.

   Gibbs Random Field has also been implemented to multi-robot systems for flocking control. In \cite{Xi2006Auto}, Xi \emph{et. al.} proposed a Gibbs sampler-based algorithm for autonomous swarms, where global objectives and constraints are encoded in Gibbs potential energies. The Gibbs sampler-based algorithm can overcome the potential problem of robots being trapped at local minima of potential/objective functions \cite{TAN20102068}. In \cite{fernando_online_2021}, the GRF-based methods are implemented to multiple quadrotors by virtue of their differential flatness property. More GRF-based designs can also be found in \cite{ Rezeck2021IROS} and the references therein. However, it should be pointed out that all the existing GRF-based designs above are not capable of pattern formation.  Recent work in \cite{rezeck2022RAL} investigates the pattern formation for robot swarms using GRF, which is inspired by chemistry rules. However, a chemistry-inspired algorithm in \cite{rezeck2022RAL} cannot form arbitrary patterns of interest, which significantly restricts its applications. The dynamic pattern formation problem is addressed in the paper based on a newly introduced exploration velocity potential that is inspired by the works in \cite{sun2023Nature}.
	
	\section{Problem formulation\label{sec:Prob}}
 
	In this paper, $N$ homogeneous robots are considered with $\mathcal{A} = \left\{1,2,..., N\right\}$ as the robot set. For each robot $i \in \mathcal{A}$, its model is given by
	\begin{equation}\label{eq:dynamics}
		\dot{\mathbf{p}}_i(t) = \mathbf{v}_i(t),\;\; \dot{\mathbf{v}}_i(t) = \mathbf{u}_i(t)
	\end{equation}
For $t\in\mathbb{R}>0$, where $\mathbf{p}_i(t),\mathbf{v}_i(t)\in\mathbb{R}^2$ are the position and velocity, $\mathbf{u}_i(t)\in\mathbb{R}^2$ is the control input. According to (\ref{eq:dynamics}), we define the robot state as $\mathbf{x}_i(t)= [{\mathbf{p}}^T_i(t), {\mathbf{v}}^T_i(t)]^T$.  
For robots $i,j \in \mathcal{A}$, $\mathbf{s}_{ij}(t) = \mathbf{p}_i(t) - \mathbf{p}_j(t)$ represents the relative position of robot $i$ to robot $j$. The neighbor set of robot $i \in \mathcal{A}$ is denoted by $\mathcal{N}_i(t) \subseteq \mathcal{A}$. In this paper, the neighbor set of robot $i$ is determined based on the k-nearest neighbors method that is also the case for gregarious animals in nature \cite{Ballerini2008PNAS}.  The safety radius for robot $i$ is denoted by $r_{coll}$ that is generally larger than the physical size of a robot.

    \textbf{\emph{Control objective:}} In the flocking system of interest, we aim to develop a distributed and efficient predictive flocking control method for the safe navigation of robots in environments with obstacles. The robot swarm is also expected to be capable of following a region-based pattern or shape.
 
	\section{Approach\label{sec:appr}}

 	\subsection{Potential energies} \label{Subsec:PotentialEnergy}
In this paper, a group of robots are expected to move safely and assemble smoothly as a flock in challenging environments. The safe motion of robots in flocking is generated by the combination of various collective behaviors of multiple robots in flocking, such as collision avoidance, obstacle avoidance, motion smoothness, and shape formation, \emph{etc.} In GRF, the aforementioned flocking behaviors are characterized in terms of a set of potential energies that are eventually used to construct the probability distribution of GRF. 
Those potential energies can also be regarded as cost functions to evaluate the flocking control performance. More details on the potential energies are given below.
	
1) \emph{Interaction potential energy}: The interaction potential energy contains a repulsive potential energy and a velocity alignment potential energy. Let $d_{ij}=\|\mathbf{s}_{ij}(t)\|$ be the relative distance of robot $i$ to robot $j$. The repulsive potential energy is, therefore, defined as
  % To simplify the calculation form, according to the method in \cite{fernando_online_2021}, the potential function $\varPsi_a$ of the interaction between robots can be selected as.
	\begin{equation}
		\psi_a(\mathbf{x}_i, \mathbf{x}_j) = \left\{
  \begin{array}{cc}
  k_{a}(1-\sin(\frac{\pi d_{ij}}{2r_{a}})), & d_{ij}<r_{a}\\
  0, & otherwise
  \end{array}
  \right.
	\end{equation}
where $k_{a}$ is a positive coefficient, $r_{a}$ is a distance threshold. If $d_{ij}\geq r_{a}$, there is no repulsion between robots $i$ and $j$.
% Out of the characteristic distance $r_a$, the minimum inter-robot potential energy can be obtained. In the absence of other interactions, the distance between two adjacent robots will be eventually kept outside $r_a$.

The alignment potential energy aims to ensure motion consistency of robots in flocking and thus reduces collision probability, which is given by
   % Although the velocity alignment principle cares more about the motion direction, the velocity magnitude also affects the energy between robots. A fast robot travels a larger distance in a given amount of time (planning horizon $t_p$), which influences other flocking rules. Hence, the velocity alignment term is chosen to be
	\begin{equation}
        \label{eq:velocity_alignment}
		\psi_{align}(\mathbf{x}_i, \mathbf{x}_j) = k_{align} \exp({d_i|\Delta \theta_{ij}|}/{k_l})
	\end{equation}
where $k_{align} > 0$, $d_i = ||\mathbf{v}_i||t_p$ with $t_p$ as the time step, and $\Delta\theta_{ij} = \arccos (\mathbf{v}_i^T \mathbf{v}_j/ \left\Vert\mathbf{v}_i\right\Vert \left\Vert\mathbf{v}_j\right\Vert)$ with $j\in \mathcal{N}_i(t)$ represents the heading difference between robots $i$ and $j$. 
 % A similar cost $\varPsi_{align}^{\beta}$ can also be defined for the improvement of obstacle avoidance, where the obstacle is taken as a $\beta$-agent as in \cite{olfati-saber_flocking_2006}. In the obstacle avoidance scenario, $\Delta\theta = \arccos (\mathbf{v}_i^T \mathbf{v}_{i\beta} / \left\Vert\mathbf{v}_i\right\Vert \left\Vert\mathbf{v}_{i\beta}\right\Vert)$, $\beta\in \mathcal{N}_i^{\beta}(t)$. 
 The interaction potential energy is thus
  \begin{equation}
      \psi_{inter}(\mathbf{x}_i, \mathbf{x}_j) = \psi_{a}(\mathbf{x}_i, \mathbf{x}_j) + \psi_{align} (\mathbf{x}_i, \mathbf{x}_j)\label{eq:InteractEnergy}
  \end{equation}
 
 %where the potential function $\varPsi_a$ reaches a minimum. The robot group, therefore, forms a fixed formation.
	
	2) \emph{Obstacle avoidance potential energy}: 
 For collision avoidance, obstacles are viewed as virtual $\beta-$agents as in \cite{olfati-saber_flocking_2006}. Hence, the interaction between robots and obstacles is 
	\begin{subequations}
		\begin{equation}
		\phi_\beta(z) = \rho_h(\frac{z}{d_\beta})(\sigma(z-d_\beta)-1)
		\end{equation}
		\begin{equation}
		\rho_h(z) = \left\{
			\begin{array}{cc}
			1, & z\in[0,h) \\
			\frac{1}{2}[1+\cos(\pi\frac{z-h}{1-h})], & z\in[h,1) \\
			0, & otherwise
			\end{array}
			\right.
		\end{equation}
	\end{subequations}
where $d_\beta$ is the radius of influential region of $\beta$, and $\sigma(z) = z / \sqrt{1+z^2}$. We use an exponential function to represent the collision avoidance potential energy.
	\begin{equation}
	\psi_o(\mathbf{x}_i,\mathbf{x}_\beta) = k_{o}\exp(-\phi_{\beta}(d_{i\beta}))
	\end{equation}
where $k_0 > 0$, $d_{i\beta} = ||\mathbf{p}_i(t) - \mathbf{p}_{\beta}(t)||$ is the distance between robot $i$ and the virtual $\beta-$agent.
Neighboring obstacles are perceived according to distance, so
	\begin{equation}
	\mathcal{N}_i^\beta(t) = \{\beta | d_{i\beta} = ||\mathbf{p}_i(t) - \mathbf{p}_{\beta}(t)|| \leq d_\beta \}
	\end{equation}
where $d_\beta>0$ is a threshold.

    \begin{figure}[!tbbp]
        \centering
        \includegraphics[width=0.9\linewidth]{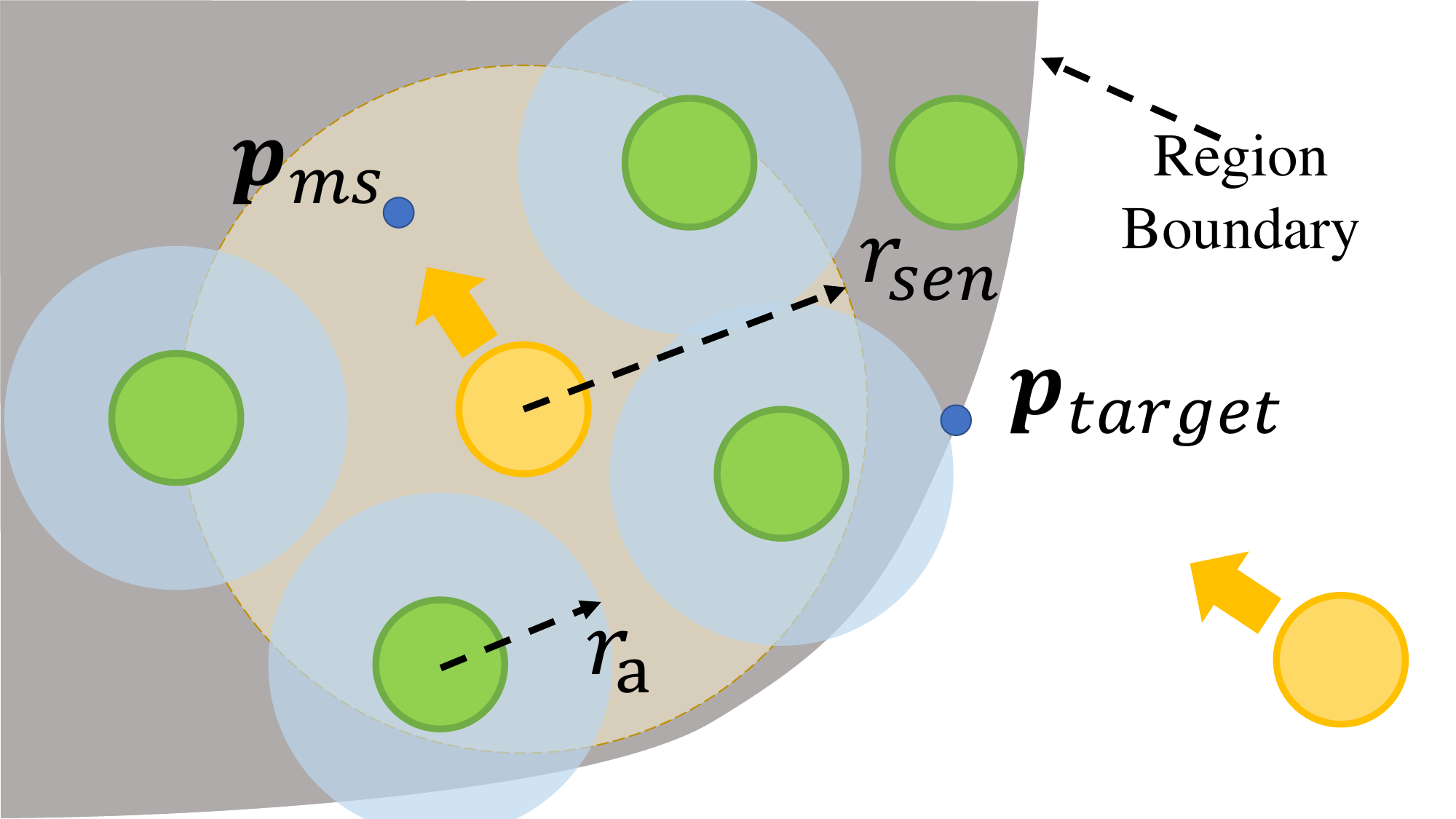}
        \caption{The mode of shape control potential energy. The yellow and green balls represent swarm robots. Robots calculate $\bold{p}_{ms}$ based on the area within $r_{sen}$ (light yellow area) that is not occupied by other robots (light blue area).}

        \label{fig:mean shift}
        \vspace{-0.5cm}
    \end{figure}
%        \begin{figure}[!htbp]
%        \centering
    
%        \includegraphics[width=\linewidth]{figures/obstacles.pdf}
%        \caption{Determination of the neighboring virtual agents. (a) A cylindrical or %circular obstacle. (b)
%        A middle point of a rectangular obstacle. (c) The apex of a rectangular obstacle.}
%        \label{fig:obstacles}
%        \vspace{-0.5cm}
%        \end{figure}
	
	It is not advisable to overemphasize distance maintenance between robots in a complex obstacle environment, as this will downgrade a robot's responsiveness to avoiding collisions. As a result, the weight of the obstacles $w_o$ is chosen as 
	\begin{equation}
		w_o = 1 - {|\phi_\beta|}/{range(\phi_\beta)}
	\end{equation}
	where $range(\phi_\beta)$ is a positive constant that specifies the difference between the upper and lower bounds of $\phi_\beta$. %\qrz{When the agent is close to obstacles, $\phi_\beta$ absolute value increases gradually, and $w_o$ gradually decreases.}
	
%Although the velocity alignment principle cares more about the motion direction, the velocity magnitude also affects the energy between robots. A fast robot travels a larger distance in a given amount of time (planning horizon $t_p$), which influences other flocking rules. Hence, the velocity alignment term is chosen to be
%	\begin{equation}
%		\varPsi_{align}(\mathbf{x}_i, \mathbf{x}_j) = k_{align} (\exp(\frac{d_i|\Delta \theta|}%{k_l}))
%	\end{equation}
%	where $d_i = ||\mathbf{v}_i||t_p$, and $\Delta\theta = \arccos (\mathbf{v}_i^T \mathbf{v}_j/ \left\Vert\mathbf{v}_i\right\Vert \left\Vert\mathbf{v}_j\right\Vert)$, $j\in \mathcal{N}_i(t)$ represents the velocity angle between robots $i$ and $j$. To better avoid the obstacle, a similar cost $\varPsi_{align}^{\beta}$ can be added between the robot and the obstacle. In this case, $\Delta\theta = \arccos (\mathbf{v}_i^T \mathbf{v}_{i\beta} / \left\Vert\mathbf{v}_i\right\Vert \left\Vert\mathbf{v}_{i\beta}\right\Vert)$, $\beta\in \mathcal{N}_i^{\beta}(t)$.
	
	% Although the influence on group performance is minimized to the greatest extent possible using \eqref{eq:2}, T
3)  \emph{Motion smoothness potential energy}: One common drawback of existing GRF-based methods is their ignorance of the motion performance of robots in flocking. It would result in oscillatory or wobbling motion behaviors that might deteriorate the system stability in real-world implementations. Hence, a motion smoothness potential energy is defined, which consists of two items $\psi_{acc}$ and $\psi_{vel}$. The first item is used to limit the size and direction change of $\mathbf{u}_i$, so one has
	\begin{equation}\label{eq:acceleration cost}
	\psi_{acc}(\mathbf{x}_i,\mathbf{u}_i) = k_{acc}(\exp(\frac{||\mathbf{u}_i||}{k_c}) + \exp(\frac{\delta\theta_i}{k_d}))
	\end{equation}
 where $k_{acc}$, $k_c$ and $k_d$ are  positive coefficients, $\delta\theta_i = \arccos(\mathbf{u}_i \cdot \mathbf{u}_{i,last} / ||\mathbf{u}_i|| \cdot ||\mathbf{u}_{i,last}||)$, and $\mathbf{u}_{i,last}$ is the previous acceleration of robot $i$. The second term is used to prevent the velocity of robot $i$ deviating too much from the desired flocking velocity $\mathbf{v}_d$, so it is given by
  \begin{equation}\label{eq:velocity cost}
        \psi_{vel}(\mathbf{x}_i) = k_{vel}(\exp({\|\mathbf{v}_i-\mathbf{v}_d\|}/{k_v})-1)
        \end{equation}
where $k_{vel}$ and $k_v$ are positive coefficients. The motion smoothness potential energy can thus be expressed as 
\begin{equation}\label{eq:MotionSmoothEnergy}
    \psi_{s}(\mathbf{x}_i,\mathbf{u}_i) = \psi_{acc}(\mathbf{x}_i,\mathbf{u}_i) + \psi_{vel}(\mathbf{x}_i)
\end{equation}

   4) \emph{Shape control potential energy}: This potential allows the swarm to form a region-based shape efficiently and evenly. It is inspired by the work in \cite{sun2023Nature}, which is quantified by using a region-attraction energy $\psi_{ro}$ and a mean shift exploration energy $\psi_{ri}$. The region of interest is divided into many small grids which is described by its relative distance from the center of region. The initial position and speed sequence of the region is sent to each robot as preset command. As shown in Fig. \ref{fig:mean shift} , when the robot $i$ is out of region, $\psi_{ro}$ can attract it toward $\mathbf{p}_{target}$ that is the position of the nearest grid to robot $i$ in the region of interest. In another case, when the robot $i$ in the region, the mean shift exploration potential energy $\psi_{ri}$ will drive robots to $\bold{p}_{ms,i}$ which represents the lowest density within region of interest for robot $i$. The region-attraction energy $\psi_{ro}$ and the mean shift exploration energy $\psi_{ri}$ is described as
    \begin{equation}
        \psi_{ro}(\mathbf{x}_i) = k_{ro}(\exp(||\mathbf{p}_i - \mathbf{p}_{target}||) - 1)
    \end{equation}
    \begin{equation}
        \psi_{ri}(\mathbf{x}_i) = k_{ri}(\exp(||\mathbf{p}_i - \mathbf{p}_{ms}||) - 1)
    \end{equation}
    where $k_{ro}, k_{ri} > 0$. In the region of interest, each robot occupies the grids within a radius of $r_a$ around it. For robot $i$, all the unoccupied grids within its sensing radius $r_{sen}$ form a set $\mathcal{M}_i$ which is used to calculate $\bold{p}_{ms,i}$.

    \begin{equation} \label{eq:msVs}
        \bold{p}_{ms,i} = \frac{ \sum_{j\in\mathcal{M}_i} f(||\mathbf{p}_i - \bold{p}_j||/r_{sen})\bold{p}_j}{\sum_{j\in\mathcal{M}_i} f(||\mathbf{p}_i - \bold{p}_j||/r_{sen})}
    \end{equation}

    \begin{equation} \label{eq:msPOS}
	f(z) = \left\{
		\begin{array}{cl}
		1, & z\in(-\infty,0] \\
		\frac{1}{2}[1+\cos(\pi z)], & z\in(0,1) \\
		0, & z\in[0,+\infty)
		\end{array}
		\right.
	\end{equation}

    Hence, the shape control potential energy is

    \begin{equation}\label{eq:ShapeControl}
        \psi_{r}(\mathbf{x}_i) = \psi_{ro}(\mathbf{x}_i) + \psi_{ri}(\mathbf{x}_i)
    \end{equation}
	
	\subsection{Gibbs random field}
	We use potential energies to describe the interactions between robots and robots and the environment, which will be used for optimizing $\mathbf{u}_i$. 
 The potential energies in Section \ref{Subsec:PotentialEnergy} are used to construct the Gibbs random field (GRF). To represent the probability distribution in the GRF, each robot is taken as a random variable at a spatial site.  Let $X = \{X_1, X_2,..., X_N\}$ represent a set of random variables of GRF, where $X_i$ corresponds to robot $i$.  The probability density $p(X)$ is factored according to the cliques of the interaction graph of robots \cite{koller2009probabilistic}. For each clique, one has
\begin{subequations}
 \begin{equation}
    \varPsi _{inter}(X_i,X_j) = \exp{(-\psi _{inter}(\mathbf{x}_i,\mathbf{x}_j))}
\end{equation}
\begin{equation}
    \varPsi _{o}(X_i,X_\beta) = \exp{(-\psi _{o}(\mathbf{x}_i,\mathbf{x}_{\beta}))}
\end{equation}
\begin{equation}
    \varPsi _{s}(X_i) = \exp{(-\psi _{s}(\mathbf{x}_i,\mathbf{u}_i))}
\end{equation}
\begin{equation}
    \varPsi _{r}(X_i) = \exp{(-\psi _{r}(\mathbf{x}_i))}
\end{equation}
\end{subequations}
where $X_{i/j/\beta}$ is the random variable corresponding to agent $i$, $j$, or $\beta$, respectively.  
The probability distribution formed by log-linear models of GRF satisfies
    \begin{equation}\label{eq:probdst}
    \begin{aligned}
     &P(X)  \propto \exp \Bigg (  -\sum _i\sum _{j \neq i} \psi _{inter}(\mathbf{x}_i,\mathbf{x}_j) -\sum _i \psi _{s}(\mathbf{x}_i,\mathbf{u}_i)  \\
     & - \sum _i \sum _{\beta\in\mathcal{N}_i^\beta(t)} \psi _{o}(\mathbf{x}_i,\mathbf{x}_\beta)- \sum _i \psi _{r}(\mathbf{x}_i)\Bigg )
    \end{aligned}
    \end{equation}
The optimal control $\mathbf{u}_i$ is obtained by maximizing an approximate distribution of $ Q(X)=\prod_i Q_i(X_i)$ at a future state $\mathbf{x}_{i, t+t_p}(\bm{u}_{i,t}, \mathbf{x}_{i,t}, t_p)$. The future state $\mathbf{x}_{i, t+t_p}(\bm{u}_{i,t}, \mathbf{x}_{i},t_p)$ is predicted based on the robot model \eqref{eq:dynamics}. Once $\mathbf{x}_{i, t+t_p}$ is obtained, $Q_i(\mathbf{x}_{i,t+t_p})$ is updated iteratively as in \eqref{eq:criterion}.
\begin{equation}\label{eq:criterion} 
	\begin{aligned}
	&Q_i(\mathbf{x}_{i, t+t_p}) = \\
 &\frac{1}{Z_i} \exp \Bigg( -w_o\sum_{j\in\mathcal{N}_i(t)} Q_j(\mathbf{x}_{j,t+t_p}) \psi_{inter}(\mathbf{x}_{i,t+t_p}, \mathbf{x}_{j,t+t_p}) \\
	&- \sum_{\beta\in\mathcal{N}_i^\beta(t)} Q_\beta(\mathbf{x}_\beta) \psi_{o} (\mathbf{x}_{i,t+t_p}, \mathbf{x}_\beta)- \psi_{s}(\mathbf{x}_{i,t+t_p},\mathbf{u}_{i,t+t_p}) \\
 & - \psi_{r}(\mathbf{x}_{i,t+t_p})\Bigg)
	\end{aligned}
	\end{equation}
where $Z_i$ is a positive normalization constant such that the total probability is equal to $1$. $Q_\beta(\mathbf{x}_\beta) = 1/\mathrm{card}(\mathcal{N}_i^\beta(t))$ satisfies a uniform distribution, indicating that obstacles encountered by robots have equal ``probability". The iteration of \eqref{eq:criterion} is initialized with $Q_i(\mathbf{x}_{i,t+t_p}) = 1 /\mathrm{card}(\mathbf{x}_{i, t+t_p}(\bm{u}_{i,t}, \mathbf{x}_{i,t}, t_p))$.
	
	\subsection{Control determination}
	% Model predictive control (MPC) is frequently employed in multiple robot motion planning and control \cite{zhan_flocking_2013,soria2021Nature}. In MPC, the current control is optimized based on current system states and the predicted forthcoming ones. Hence, MPC-based methods can potentially make proactive decisions. Inspired by the proactive feature of MPC, we also employ predictions in our flocking control. 
The optimal control input $\mathbf{u}_i$ is screened out by maximizing a posterior distribution of GRF, which would minimize the total potential energies given in Section \ref{Subsec:PotentialEnergy}. The original optimization problem is nonlinear and intractable for online implementation. To resolve the problem, the control input space is discretized uniformly as in the dynamic window approach \cite{Fox1997}. Through input discretization, the computational cost of the optimization problem can be reduced to an acceptable size at the price of sub-optimality.
	
The input space is firstly discretized into zero and non-zero cases. For the non-zero cases, the input space is discretized uniformly in the plane with the following direction interval.
 % In \qrz{The robot state is classified as either uniform velocity or acceleration. The former has an acceleration of $\mathbf{u}=0$, and the latter is assumed to be uniformly distributed in the plane.
	\begin{equation}\label{eq:2}
	\theta_{min} = {2\pi}/{n_a}
	\end{equation}
	where $n_a$ represents the number of nonzero acceleration control input directions, $\theta_{min}$ is the angle between discretized adjacent non-zero control inputs. Hence, the input discretization for robot $i$ is denoted as 
 \begin{equation}\label{eq:discretizedInput}
    \mathbf{u}_{i,t}\in\left\{\mathbf{0},{u}_{i,t}\mathbf{e}_u(\theta_{min}), \ldots, {u}_{i,t}\mathbf{e}_u(n_a\theta_{min})\right\}
 \end{equation}
 where $\mathbf{e}_u(x) = \left[\sin(x),\cos(x)\right]^T$, and ${u}_{i,t}$ is selected as an arithmetic progression with the interval $\bigtriangleup u$. Based on \eqref{eq:discretizedInput} and Eq. \eqref{eq:dynamics}, a set of future states are predicted, which will be used to generate a set of posterior distributions of GRF. The ``best'' input is selected as the one with the maximum a posteriori (MAP) in the whole distribution set.
 
 To ensure the feasibility of the planned trajectory, it is necessary to explicitly impose constraints that limit the maximum control input and speed of each individual robot. That is, for $i \in \mathcal{A}$, $||\mathbf{v}_i(t)|| \leq v_{max}$ and $||\mathbf{u}_i(t)|| \leq u_{max}$. Additionally, a self-propelled particle model is considered in the design, which is inconsistent with actual physical systems. As a consequence, low-pass filtering is introduced as below.
	\begin{equation}\label{eq:filter}
		\mathbf{u}_{i} = (1 - \alpha)\mathbf{u}_{last,i} + \alpha\mathbf{u}_i^*
	\end{equation}
        where $\alpha$ is a positive number less than 1, and $\mathbf{u}_i^*$ is the ``best'' input with the MAP according to \eqref{eq:criterion}. Reducing $\alpha$ slightly will have the same effect as \eqref{eq:acceleration cost}.
 % In addition, we specify an equal number of control inputs in both directions.
 
	% Based on \eqref{eq:dynamics}, the future states can be predicted for a control input selection $\mathbf{u}_t$.
	% \begin{equation}\label{eq:3}
	% \mathbf{x}_{t+t_p} = A\mathbf{x}_t + B\mathbf{u}_t
	% \end{equation}
	% where $\mathbf{x}_t$ and $t_p$ represent the current state and planning horizon, respectively, $A$ and $B$ are given as
	% \begin{equation}
	% 	A = \left[
	% 	\begin{array}{cc}
	% 	1 & t_p \\
	% 	0 & 1
	% 	\end{array}
	% 	\right], B = \left[
	% 	\begin{array}{c}
	% 	t_p^2 / 2 \\ t_p
	% 	\end{array}
	% 	\right]
	% \end{equation}
 
	%  Hence, predicted states at $t_p$ are $\mathbf{x}_{t+t_p}(\bm{\mu},\mathbf{x}_t,t_p)=[\mathbf{x}_{t+t_p,1}^T(\mathbf{u}_1,\mathbf{x}_t,t_p),\mathbf{x}_{t+t_p,2}^T(\mathbf{u}_2,\mathbf{x}_t,t_p),\ldots]^T$. 
  
  % The planning horizon is different from the actual control horizon $t_c$ which is the step size $t_{step}$. Instead, $t_p \geq t_c$ is generally used as an insurance policy to achieve better-flocking performance. 
  % \begin{equation*}
  %    \mathbf{x}_{t+t_p}(\bm{\mu},\mathbf{x}_t,t_p) =[\mathbf{x}_{t+t_p,1}^T(\mathbf{u}_1,\mathbf{x}_t,t_p),\mathbf{x}_{t+t_p,2}^T(\mathbf{u}_2,\mathbf{x}_t,t_p),\ldots]^T 
  % \end{equation*}

	% \subsection{Smoothing}

	\section{Simulation and experiments}\label{sec:exp}
    The flocking performance is evaluated by an order metric, a distance metric and an obstacle-avoidance metric.
	% We use the following metrics to evaluate the robot flocking performance when traversing a narrow space with obstacles.
	
 The \emph{order metric} evaluates the motion alignment performance, which is computed by 
	\begin{equation}
		order = \frac{1}{N} \sum_{i \in \mathcal{A}} \frac{1}{N_i - 1}\sum_{j \in \mathcal{N}_i(t)} \frac{\mathbf{v}_i \cdot \mathbf{v}_j}{||\mathbf{v}_i|| \cdot ||\mathbf{v}_j||}
	\end{equation}
where $N_i=\mathrm{card}(\mathcal{N}_i(t)) + 1$. The closer the order is to 1, the better the consistency of the flocking.

The \emph{distance metric} evaluates the relative distance maintenance performance, which is calculated according to the distance between each robot and its nearest neighbor, so
 \begin{equation}\label{eq:distance}
	\begin{aligned}
		d_i^{min} &= \min\left\{d_{ij} \vert d_{ij}=\|\mathbf{s}_{ij}\|, \; \forall j\in\mathcal{A}, j\neq i\right\} \\
  d^{min} &= \min\left\{d_{i}^{min} \vert \; \forall i\in\mathcal{A}\right\} \\
     d^{max} &= \max\left\{d_{i}^{min} \vert \; \forall i\in\mathcal{A}\right\} \\
    d^{avg} &= \mathrm{mean}\left\{d_{i}^{min} \vert \; \forall i\in\mathcal{A}\right\} \\
	\end{aligned}
	\end{equation}
 where the minimum, maximum, and average distance among all robots are also provided.
	
 The \emph{obstacle-avoidance metric} $d_{\beta}^{min}$ evaluates the obstacle avoidance performance, which is 
\begin{equation}\label{eq:obstacle}
	d_{\beta}^{min} = \min_{i \in \mathcal{A}, \beta \in \mathcal{N}_i^\beta(t)} ||\mathbf{p}_i(t) - \mathbf{p}_{\beta}(t)||
	\end{equation}
  where the $d^{min}$ and $d_{\beta}^{min}$ should be guaranteed to meet safety constraints. A larger $d^{min}$ and $d_{\beta}^{min}$ implies more safety.
    \begin{table}[bp]
          \vspace{-0.3cm}
	    \centering
            \caption{Algorithm parameter setups}
            \setlength{\tabcolsep}{0.6em}{
	    \begin{tabular}{cccc}
            \toprule
              Parameters  & Simulation $1$ & Simulation $2$ & Real experiment \\
            \midrule
              $k_a$ & 0.8 & \textbf{0.7} & \textbf{0.7} \\
              $k_o$ & 1 & 1 & 1 \\
              $k_l$ & 4 & 4 & 4 \\
              $k_c$ & 7 & 7 & 7 \\
              $k_d$ & 15 & 15 & 15 \\
              $k_v$ & 2 & 2 & 2 \\
              $k_{align}$ & 0.2 & 0.2 & 0.2 \\
              $k_{acc}$ & 10 & 10 & 10 \\
              $k_{vel}$ & 0.07 & 0.07 & 0.07 \\
              $k_{ro}$ & 5 & \textbf{25} & \textbf{25} \\
              $k_{ri}$ & 10 & 10 & 10 \\
              $h$ & 1/3 & 1/3 & 1/3 \\
              $d_{\beta}$ & 0.2 & 0.2 & 0.2 \\
              $n_a$ & 6 & 6 & 6 \\
              $\bigtriangleup u$ & 1/6 & 1/6 & 1/6 \\
              $t_p$ & 0.15 & \textbf{0.15} & \textbf{0.2} \\
              $\alpha$ & 0.9 & 0.9 & 0.9 \\
              $r_{f}$ & 0.3 & \textbf{0.3} & \textbf{0.4} \\
              $r_{sen}$ & 0.5 & 0.5 & 0.5 \\
            \bottomrule
	    \end{tabular}
        }
	   
	    \label{tab:our method's parameters}
     % \vspace{-0.25cm}
	\end{table}
  
    % where $t_{step}$ is the step size at simulation and experiments.
%	The \textbf{control efficiency metric} evaluates the input efficiency of each robot, which is defined as  
 % Maintaining a higher value of acceleration requires more resources, which is not desired. Each agent's average acceleration should be measured.
%	\begin{equation}\label{eq:mean control input}
%	u_i^{avg} = \frac{1}{T} \sum_{t=0}^{T} ||\mathbf{u}_i(t)||
%	\end{equation}
% where $T$ is the total running time.
 
%	The \textbf{trajectory length metric} evaluates the distance traveled by robots. A short trajectory length is  preferable in general.
% \begin{equation}\label{eq:trajectory length}
%	L_i = \sum_{\substack{t=0 }} ^{T}||\mathbf{p}_i(t+t_{step}) - \mathbf{p}_i(t)||
%	\end{equation}
 
	% \begin{equation}\label{eq:trajectory length}
	% L_i = \sum_{\substack{t, t_{last} \leq T \\t - t_{last} = t_c}} ||\mathbf{p}_i(t) - \mathbf{p}_i(t_{last})||
	% \end{equation}
 
    %\begin{figure}[htbp]
        %\centering
        %\subfigure[]{
            %\label{fig:complex scenario }
            %\includegraphics[width=0.48\linewidth]{figures/complex scenario.pdf}
        %}
        %\subfigure[]{
            %\label{fig:simple scenario}
            %\includegraphics[width=0.32\linewidth]{figures/simple scenario.pdf}
        %}
       % \includegraphics[width=0.7\linewidth]{figures/scene.pdf}
      %  \caption{The proposed algorithm was tested using a narrow passage scenario. In the map, blue dots represent navigation points, blue arrows represent the group's expected path, and orange areas represent obstacles.}
      %  \label{fig:scenarios}
       % \end{figure}

 \begin{figure}[tbp]
        \centering
    \includegraphics[width=0.98\linewidth]{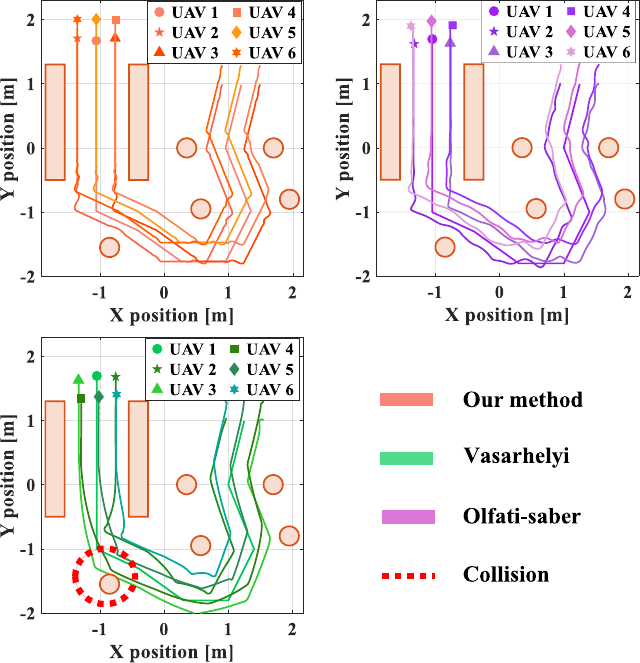}
        \caption{Flocking trajectories (Simulation 1).}
        %The Olfati-saber's is slightly oscillating, and the Vásárhelyi's collides due to insufficient reaction ability.}
        \label{fig:simulation trajectory}
       \vspace{-0.5cm}
    \end{figure} 
\begin{figure}[bp]
	\centering	
 \vspace{-0.3cm}
 \includegraphics[width=0.95\linewidth]{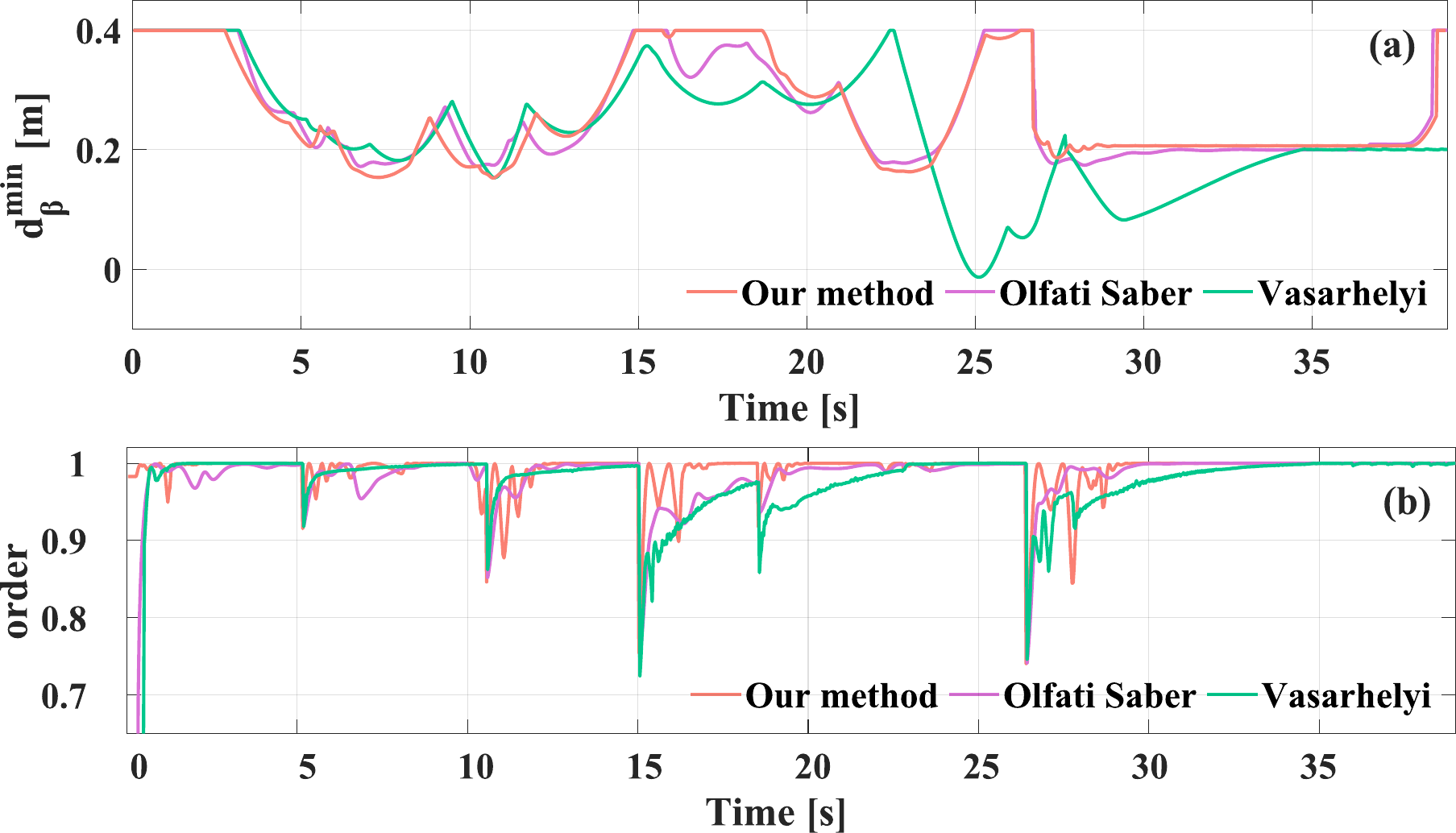}
        \caption{Flocking performance (Simulation 1). (a) denotes the distance between the robot and the obstacle. (b) shows the croup consistency (order).}

        \label{fig:metric}
\end{figure}

	\subsection{Simulation}
    We conducted two simulations to evaluate the proposed method in different scenarios. In the first simulation scenario, the flock with 6 robots moves along a planned route to pass a narrow space. We also compared our method to Olfati-Saber's method \cite{olfati-saber_flocking_2006} and V\'{a}s\'{a}rhely's method \cite{Vasarhelyi2018SR} to prove that our method has better performance in collision avoidance, trajectory smoothing and maintain formation.
    %The simulation scenario is depicted in Fig. \ref{fig:scenarios}. 
    In the second simulation, a butterfly pattern is designed to validate our method's ability to pattern formation in an obstacle-free scene. The butterfly has a size of $4.4$ meter and is divided into $9752$ grids. The shape moves along an S-curve at a speed of $0.12 \mathrm{m/s}$. Initially, 50 robots are placed randomly. Each robot is capable of detecting the states of the 3 closest robots. The maximum acceleration and velocity of the robot are limited to $a_{max} = 0.7$ $\mathrm{m/s^2}$ and $v_{max} = 0.4$ $\mathrm{m/s}$. The key algorithm parameters corresponding to the two simulations are shown in the first two columns of Table \ref{tab:our method's parameters}. In the first simulation, we pay more attention to the collision-avoidance performance of the flocking, so a larger $k_a$ is selected. In other parts, we emphasize the shape assembly performance of our method, so a larger $k_{ro}$ is a better choice.

    The results of the first simulation are shown in Fig. \ref{fig:simulation trajectory} and Fig. \ref{fig:metric}.
    %both our method and Olfati-saber's method can achieve safe motion in the  cluttered simulation scenario. The Vásárhelyi method cannot avoid collision with some obstacles. Collision with obstacles occurs to the 
    %%%%%%%%%%%%%%%%%%%
    %%%%%%%%%%%%%%%%%%%poor formation
    The Olfati-Saber's method has a twisted trajectory because the motion performance of robots in flocking is ignored. The Vásárhelyi's method lacks a consistent shape when passing narrow space, and cannot guarantee basic security. These two methods are pre-designed with insufficient responsiveness when executing. They calculate the acceleration command by directly adding the different effects.
    Our method predicts flocking control input by considering the interaction of multiple potential energies and designs weight, such as $w_o$, to achieve better performance on dynamically adapting to complex environment.

    as shown in Fig. \ref{fig:metric} (b), all of the three methods can ensure converged group consistency in an open environment, but our method has the largest convergence rate.
     \begin{figure}[tbp]
            \centering
            \includegraphics[width=1\linewidth]{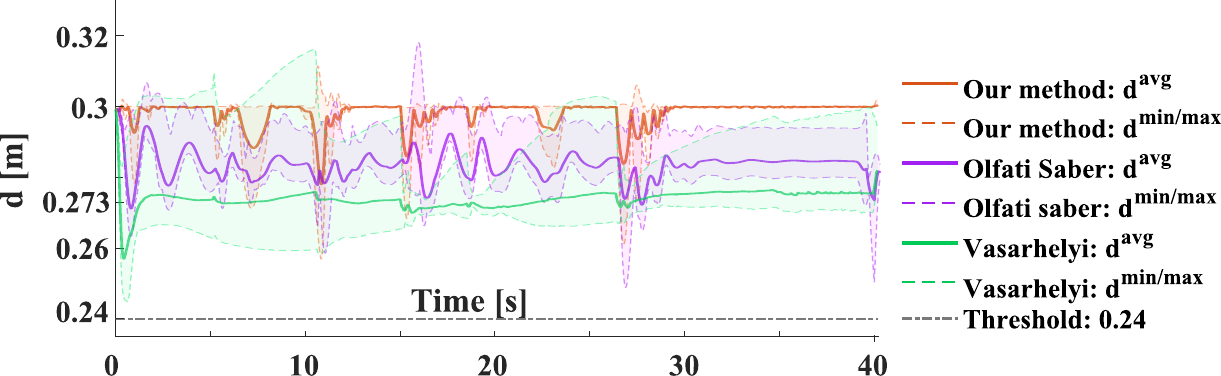}
            \caption{Distance metric (Simulation 1). Areas covered by light colors represent the variation of the metric.}
            \label{fig:distance}
             \vspace{-0.5cm}
        \end{figure}

    By comparing the light-colored areas between $d^{min}$ and $d^{max}$ in Fig. \ref{fig:distance}, our method performs better in maintaining distance between the nearest robots. The robots in flocking will quickly return to a stable state despite fluctuations at corners or obstacles. That's because our method introduces shape control potential energy, its combined effect with the repulsive potential energy allows robots to maintain relatively stable.
    The Olfati-Saber's method has a large oscillation to maintain group configuration. The ability of the Vásárhelyi's method to retain configuration is limited.

\begin{figure}[tbp]
 % \vspace{-0.3cm}
    \centering           \includegraphics[width=0.95\linewidth]{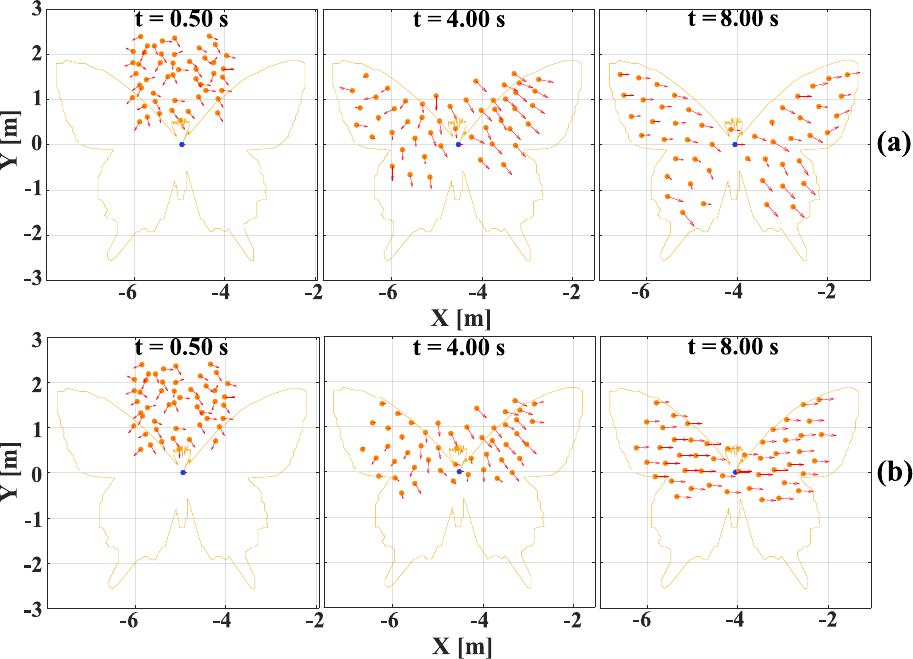}
    \caption{Pattern formation. Red arrows are the heading angles of robots. (a) with mean shift. (b) without mean shift.}
    \label{fig:snapshots}
    \vspace{-0.3cm}
\end{figure}

\begin{figure}[tbp]
    \centering
    \includegraphics[width=0.9\linewidth]{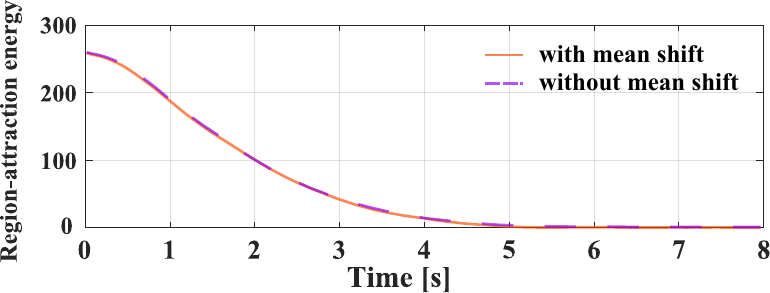}
    \caption{The region-attraction energy.}
    \label{fig:potential}
    \vspace{-0.4cm}
\end{figure}

    The results of the second simulation are reported in Fig. \ref{fig:snapshots} and Fig. \ref{fig:potential}. After $t=5\ \mathrm{s}$, the region-attraction energy reaches $0$ in both cases (with and without mean shift), so all robots converge to the region of interest. However, the algorithm without mean shift fails to allow robots to spread evenly in the region of interest. Hence, the region-attraction energy drives robots towards the region, while mean shift is mainly used to form the desired pattern in our design.

	\subsection{Experiments}
 Experiments are also performed to validate the proposed algorithm in real implementations. In the experiments, the algorithm is implemented on ten DJI Tello quadrotors.
    % To demonstrate the practical feasibility, we also put the proposed method through its paces with 4 DJI Tello drones. 
    %The experiment has the similar environment setup as the complex simulation scenario shown  in Fig. \ref{fig:real trajectory} (a).
    % \ref{fig:real snapshots}, route selection is less complicated than the simulation's complex scenario. 
    % The states of Tello quadrotors are measured by OptiTrack optical motion capture system, as shown in Fig. \ref{fig:real communication}.
     The states of Tello drones are measured by OptiTrack optical motion capture system, which
    could transmit Tello's position and velocity information at 120 Hz.
    We designed three patterns of "S", "Y" and "U". The quadrotors are initially distributed with $0.5\  \mathrm{m}$ spacing on a line. As shown in Fig. \ref{fig:real trajectory}, drones can form "S", "Y", "S", "U" patterns in a sequence. 

\begin{figure}[htbp]
            \centering
\includegraphics[width=1\linewidth]{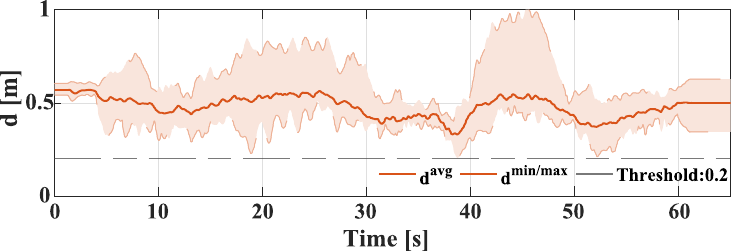}
           \caption{Distance metric (Experiments). Areas covered by light colors represent the variation of the metric.}
            \label{fig:realexp_dis}
            \vspace{-0.25cm}
        \end{figure}

    Tello drones can only track velocity commands. They do not provide the interface for the acceptance of acceleration commands. Hence, we modified the acceleration commands to the velocity ones as below.
        \begin{equation}\label{eq:velocity command}
            \mathbf{v}_i^*=\mathbf{v}_i+\mathbf{u}_i^* t_{p}
        \end{equation}
Snapshots of the experiments are given in Fig. \ref{fig:real trajectory}. The distance metric shown in Fig. \ref{fig:realexp_dis} illustrates there is no collision during the formation process. Although the drones used in the experiment have a very different dynamic model compared to the first-order dynamic model, our method performs well in collision avoidance and dynamic pattern formation.

	\section{Conclusions} \label{sec:Conclusion}

 This paper focuses on a method for implementing the flocking movement in a complex environment. Using the strategy of the potential function, we focus on the problems of obstacle influence, trajectory smoothness, and shape formation. From Gibbs random field, we converted the problem of selecting the predicted states into the problem of calculating the joint probability distribution. Compared with two existing bio-inspired methods, our method performed better in many aspects. A real experiment was also conducted to validate the assembly capability.

 Further research includes extending the method to arbitrary dimensions and conducting a larger-scale flocking control study as well as an outdoor experiment.

	%%%%%%%%%%%%%%%%%%%%%%%%%%%%%%%%%%%%%%%%%%%%%%%%%%%%%%%%%%%%%%%%%%%%%%%%%%%%%%%%

	%%%%%%%%%%%%%%%%%%%%%%%%%%%%%%%%%%%%%%%%%%%%%%%%%%%%%%%%%%%%%%%%%%%%%%%%%%%%%%%%

	%%%%%%%%%%%%%%%%%%%%%%%%%%%%%%%%%%%%%%%%%%%%%%%%%%%%%%%%%%%%%%%%%%%%%%%%%%%%%%%%

	\bibliography{myRefs.bib}
	\bibliographystyle{IEEEtran}
	
\end{document}